\pdfoutput=1
\documentclass[11pt]{article}

\usepackage{emnlp2023}

\usepackage{times}
\usepackage{latexsym}
\usepackage[T1]{fontenc}
\usepackage[utf8]{inputenc}
\usepackage{microtype}
\usepackage{inconsolata}

\usepackage{amsmath,amssymb,amsthm}  % \mathbb (lines 84, 108, 830+), \qed
\usepackage{graphicx}                % \includegraphics, \resizebox
\usepackage{booktabs}                % \toprule/\midrule/\bottomrule (line 1008+)
\usepackage{float}
\usepackage{algorithm}
\usepackage{algpseudocode}

\usepackage{pifont}
\newcommand{\cmark}{\ding{51}}
\newcommand{\xmark}{\ding{55}}

\usepackage{hyperref}

\title{Mitigating Demographic Disparities in Explanation--Human Alignment via Rationale Regularization}

\author{Suryansh Talukdar, Taha Sarwat, and Jonathan Rusert \\
        Purdue University Fort Wayne \\
        \texttt{\{talusb01, sarwt01, jrusert\}@pfw.edu}}

\begin{document}
\maketitle
\begin{abstract}
We study explanation fairness in hate speech detection, where strong predictive performance does not guarantee equally plausible explanations across demographic groups. Using the HateXplain corpus, we introduce Plausibility-EFG, a group-level disparity metric that measures the worst-case gap in explanation plausibility via alignment between token attributions and human rationale annotations. We evaluate BERT, RoBERTa, and an adversarially trained variant, ADV-RoBERTa, using Integrated Gradients and SHAP. Results show that ADV-RoBERTa improves both classification performance and rationale alignment while substantially reducing subgroup explanation disparities under both attribution methods. These improvements remain consistent across demographic identities and stable across train, validation, and test splits. We further show that prediction-level debiasing alone does not eliminate explanation inequity, indicating that predictive fairness and explanation fairness are related but distinct objectives. Finally, fairness–performance ablations reveal a controlled tradeoff in which stronger adversarial regularization reduces Plausibility-EFG at the cost of modest macro-F1 degradation. Overall, the paper establishes explanation fairness as a measurable objective in hate speech moderation and presents Fair-REF as an effective mitigation strategy for reducing subgroup disparities in model explanations.

\end{abstract}

\section{Introduction}
Hate speech moderation is a high-stakes setting in which transformer classifiers are often evaluated primarily by predictive performance, even though their token-level explanations are used to support reviewer trust and borderline moderation decisions. This makes explanation quality practically consequential, not merely diagnostic. Recent work has already shown that explanation behavior can vary across demographic proxies in hate speech settings, and that such disparities may persist even when classification looks strong \cite{mathew2021hatexplain}, \cite{balagopalan2022road}, \cite{brandl2024interplay}. Consequently, predictive fairness does not imply explanation fairness, and explanation quality should be audited independently of output-level performance.

This paper studies a methodologically underexplored failure mode: explanations can be less plausible for some demographic groups than others. We introduce Plausibility-EFG, a worst-case group disparity metric for explanation--human alignment on HateXplain, where plausibility is measured by AUPRC over attribution scores and human rationales. This follows the faithfulness--plausibility distinction \cite{jacovi2020towards,deyoung2020eraser} and tests the orthogonality hypothesis that prediction-level fairness does not necessarily imply explanation-level fairness \cite{balagopalan2022road,brandl2024interplay}. The paper contributes Plausibility-EFG, the first subgroup rationale benchmark for HateXplain, a unified evaluation framework, and Fair-REF, a mitigation pipeline that improves explanation fairness while preserving strong task performance \cite{sundararajan2017axiomatic,lundberg2017unified}.

\section{Related Work}

Post-hoc explanation methods are now standard tools for analyzing neural NLP classifiers, especially in settings where token-level evidence is important. Gradient-based methods such as Integrated Gradients provide axiomatic attribution scores by integrating gradients from a baseline to the input and have become a common choice for transformer interpretability \cite{sundararajan2017axiomatic}. SHAP offers a complementary, game-theoretic view of feature attribution by estimating token contributions through Shapley values \cite{lundberg2017unified}. In NLP, these methods are frequently paired with rationale benchmarks such as ERASER and HateXplain, where model attributions can be compared against human-annotated evidence spans \cite{deyoung2020eraser}, \cite{mathew2021hatexplain}. Rationale extraction models such as REF further move explanation generation into the training objective by jointly optimizing prediction and sparse token selection \cite{lei2016rationalizing}.

A key distinction in explanation evaluation is between faithfulness and plausibility: faithfulness asks whether an explanation reflects the model’s decision process, while plausibility asks whether it matches human judgment or annotation. These criteria are not interchangeable and should be evaluated separately; in rationale-based NLP, token overlap with human evidence provides a natural plausibility signal \cite{jacovi2020towards,deyoung2020eraser}. Fairness research in NLP has mostly focused on prediction-level disparities and label-level mitigation, but recent work shows that explanation quality can also differ across demographic groups. In particular, explanation fidelity varies across protected subgroups in non-NLP settings, and fairness and explainability do not necessarily co-improve in NLP, suggesting that explanation fairness is a distinct objective \cite{blodgett2020language,zhang2018mitigating,balagopalan2022road,brandl2024interplay}.

Our work differs from prior explanation-fairness studies in several ways. First, we study plausibility rather than fidelity, asking whether attributions align with human rationales rather than approximate the model’s internal logic \cite{balagopalan2022road}. Second, we use HateXplain, an NLP-native benchmark with token-level rationales and demographic target labels, to measure explanation--human alignment fairness directly \cite{mathew2021hatexplain}. Third, we go beyond auditing by proposing Fair-REF, a training-time mitigation objective that regularizes group-level rationale quality. This extends prior audits from diagnosis to mitigation and reflects the emerging view that prediction fairness and explanation fairness can be orthogonal \cite{brandl2024interplay}. Related work on fair explanations in moderation further shows that uneven explanation quality can have disparate impacts on content moderators, motivating our orthogonality hypothesis \cite{nguyen2023towards}.

\section{Theoretical Justification}

Let $D=\{(x_i, y_i, g_i, r_i)\}_{i=1}^N$ denote a corpus of posts, class labels, demographic groups, and human rationale masks, where $x_i=(x_{i1}, \ldots, x_{iT_i})$ is a token sequence, $y_i \in \{1, \ldots, C\}$, $g_i \in G$, and $r_i \in \{0, 1\}^{T_i}$. The classifier $f_\theta$ induces logits $z_\theta(x_i) \in \mathbb{R}^C$ and predictive distribution $p_\theta(y \mid x_i) = \mathrm{softmax}(z_\theta(x_i))$. The central hypothesis of this work is that hate-speech modeling should be optimized not only for predictive accuracy, but also for rationale alignment and subgroup-consistent explanation quality, which is well motivated by rationale-based NLP, benchmarked explanation evaluation, and recent work on fairness as decision-rationale alignment \cite{lei2016rationalizing, deyoung2020eraser, mathew2021hatexplain}.

We therefore define the training objective as a three-term multi-objective program:
\begin{equation*}
L_{\text{total}}(\theta, \phi) = L_{\text{task}}(\theta) + \lambda_{\text{rat}} L_{\text{rat}}(\theta) + \lambda_{\text{adv}} L_{\text{adv}}(\theta, \phi),
\end{equation*}
with supervised task loss
\[
L_{\text{task}}(\theta) = -\frac{1}{N} \sum_{i=1}^N \log p_\theta(y_i \mid x_i),
\]
rationale-alignment loss
\[
L_{\text{rat}}(\theta) = \frac{1}{N} \sum_{i=1}^N \ell_{\text{align}}(s_i, r_i),
\]
and adversarial demographic loss
\[
L_{\text{adv}}(\theta, \phi) = -\frac{1}{N} \sum_{i=1}^N \log q_\phi(g_i \mid h_\theta(x_i)).
\]
Here $h_\theta(x_i)$ is a shared representation and $q_\phi$ is the group discriminator (adversary). Under gradient reversal, the effective optimization is 
\[
\min_\theta \max_\phi \; L_{\text{task}}(\theta) + \lambda_{\text{rat}} L_{\text{rat}}(\theta) - \lambda_{\text{adv}} L_{\text{adv}}(\theta, \phi),
\]
which follows the standard adversarial debiasing formulation of \cite{zhang2018mitigating} and the rationale-alignment framing of \cite{li2023fairer}.

For post hoc explanation, let $A_\theta(x_i) \in \mathbb{R}^{T_i}$ denote a token attribution operator and let $s_i = \mathrm{Norm}(\max(A_\theta(x_i), 0)) \in [0, 1]^{T_i}$ be the normalized non-negative attribution vector. In the case of Integrated Gradients, the attribution for token $t$ is
\[
\mathrm{IG}_t(x) = (x_t - x'_t) \int_0^1 \frac{\partial F(x' + \alpha(x - x'))}{\partial x_t} \; d\alpha,
\]
with numerical approximation
\[
\mathrm{IG}_t(x) \approx (x_t - x'_t) \cdot \frac{1}{m} \sum_{k=1}^m \frac{\partial F\left(x' + \frac{k}{m}(x - x')\right)}{\partial x_t},
\]
which follows the axiomatic attribution framework of \cite{sundararajan2017axiomatic}. For SHAP, the contribution of token $t$ is defined by the Shapley-value estimator
\[
\phi_t = \sum_{S \subseteq N \setminus \{t\}} \frac{|S|! (|N| - |S| - 1)!}{|N|!} \bigl(f(S \cup \{t\}) - f(S)\bigr),
\]
following \cite{lundberg2017unified}. Using both methods is theoretically important because gradient-path and coalition-based explainers can preserve model ordering while differing in attribution geometry and scale, a distinction that matters for fairness analysis \cite{dai2022fairness, brandl2024interplay}.

Given attribution vector $s_i$ and rationale mask $r_i$, plausibility is evaluated by ranking-based and overlap-based criteria:
\begin{equation}
\mathrm{AUPRC}(s_i,r_i)
=\int_0^1 P_i(\tau)\, dR_i(\tau).
\end{equation}

where $\tau_i=\mathrm{median}(s_i)$ and
$\hat r_{it}=\mathbf{1}[s_{it}\ge\tau_i]$ denotes the
binarized rationale prediction.
\[
\mathrm{IOU\text{-}F1}(i) = \frac{|\hat r_i \cap r_i|}{|\hat r_i \cup r_i|}.
\]
This follows the standard evidence-plausibility setup used in ERASER and HateXplain, where human rationales are treated as evaluative evidence spans rather than absolute causal ground truth \cite{deyoung2020eraser, mathew2021hatexplain, jacovi2020towards}.

To quantify subgroup fairness of explanations, let $D_g = \{i : g_i = g\}$ denote the subgroup-specific evaluation set and let $M_i \in \{\mathrm{AUPRC}(s_i, r_i), \mathrm{IOU\text{-}F1}(i)\}$. The group-wise mean plausibility is
\[
\mu_g = \frac{1}{|D_g|} \sum_{i \in D_g} M_i.
\]
We define the explanation fairness gap as the worst-case inter-group disparity
\[
\mathrm{EFG} = \max_{g,g' \in G} |\mu_g - \mu_{g'}|.
\]
This is the central fairness quantity in the paper. It directly operationalizes the claim that explanations should be comparably plausible across demographic groups, and it is conceptually aligned with recent work showing that explanation quality can itself be unevenly distributed across subpopulations \cite{dai2022fairness, balagopalan2022road, nguyen2023towards, brandl2024interplay}.

We further assess statistical heterogeneity across groups by the Kruskal–Wallis statistic
\[
H = \frac{12}{N(N + 1)} \sum_{g=1}^{|G|} \frac{R_g^2}{n_g} - 3(N + 1),
\]
where $R_g$ is the sum of ranks of subgroup $g$ and $n_g = |D_g|$, and practical disparity by Cohen’s effect size between the best and worst groups,
\[
d = \frac{\mu_{g^\star} - \mu_{g^\dagger}}{s_{\text{pooled}}}, \quad s_{\text{pooled}} = \sqrt{\frac{s_{g^\star}^2 + s_{g^\dagger}^2}{2}},
\]
where $g^\star$ and $g^\dagger$ denote the best- and worst-performing demographic groups, respectively. The combination of EFG, significance testing, and effect size separates worst-case gap magnitude from rank-level distributional differences, which is important because subgroup imbalance can distort simple point estimates \cite{dai2022fairness, balkir2022challenges}.

The main theoretical claim of this work is that prediction fairness and explanation fairness are only partially aligned. In gradient form, this can be expressed as the fact that task gradients $\nabla_\theta L_{\text{task}}(\theta)$ and fairness-oriented gradients $\nabla_\theta \mathrm{EFG}(\theta)$ need not be parallel or positively proportional,
\[
\nabla_\theta L_{\text{task}}(\theta) \neq c \,\nabla_\theta \mathrm{EFG}(\theta)
\quad \text{for any fixed } c > 0,
\]
so that improving classification risk or even output-level fairness does not guarantee improved explanation equity. This observation is consistent with recent analyses of the interplay between fairness and explainability in NLP and with explanation-quality disparity studies showing that post hoc explanations may remain biased even for strong models \cite{dai2022fairness, balagopalan2022road, balkir2022challenges, brandl2024interplay}.

Finally, we interpret Fair-REF as a conceptually constrained multi-objective program:
\[
\min_\theta \; L_{\text{task}}(\theta) + \lambda_{\text{rat}} L_{\text{rat}}(\theta) \quad \text{s.t.} \quad \mathrm{EFG}(\theta) \le \varepsilon.
\]
Because $\mathrm{EFG}$ is non-differentiable and attribution-dependent, we do not optimize it directly, but instead introduce $L_{\text{adv}}$ as a representation-level surrogate that suppresses linearly decodable demographic information with the empirical aim of reducing attribution-level disparities. The implemented objective can be written as
\[
J(\theta, \phi, \lambda_{\text{adv}}) = L_{\text{task}}(\theta) + \lambda_{\text{rat}} L_{\text{rat}}(\theta) + \lambda_{\text{adv}} L_{\text{adv}}(\theta, \phi).
\]
This framing makes the fairness–utility tradeoff explicit and explains why the $\lambda_{\text{adv}}$-sweep induces a Pareto frontier rather than a monotone improvement in every metric: increasing adversarial pressure generally reduces EFG but can moderately degrade Macro-F1, as observed empirically in our fairness ablations. The analytical properties of EFG, the variance behavior of subgroup estimators.

\subsection*{Core Analytical Statements}

\textbf{Proposition 1 (Basic properties of EFG).} For any set of group means $\{\mu_g\}_{g \in G} \subset [0, 1]$, $\mathrm{EFG} \in [0, 1]$, is symmetric in the group indices, and equals zero if and only if $\mu_g = \mu_{g'}$ for all $g, g' \in G$. Proof in Appendix C.1.

\textbf{Proposition 2 (Subgroup-scarcity variance effect).} If $M_i$ has finite variance within each group $g$, then $\mathrm{Var}(\hat \mu_g) = \sigma_g^2 / |D_g|$; hence low-support groups induce higher uncertainty in $\hat \mu_g$, $\mathrm{EFG}$, and associated effect-size estimates. Proof in Appendix C.2.

\textbf{Observation 1 (Partial orthogonality; empirical).} Reducing $L_{\text{task}}$ or increasing predictive performance does not in general imply a reduction in $\mathrm{EFG}$. The empirical evidence is that RoBERTa improves classification over BERT while subgroup gaps remain non-trivial, and explanation fairness improves substantially only under ADV-RoBERTa with Fair-REF. A sufficient explanation is that task gradients and explanation-fairness gradients need not be aligned in parameter space; Appendix C.3 provides an intuitive argument, but we treat this claim as primarily empirical.

\textbf{Proposition 3 (Conditional tradeoff under fairness regularization).} For sufficiently large $\lambda_{\text{adv}}$ on data distributions where demographic attributes correlate with task-relevant features, minimization of $J(\theta, \phi, \lambda_{\text{adv}})$ can reduce representation-level group information while also suppressing some task-relevant signals, yielding a non-trivial fairness–utility Pareto tradeoff. Proof sketch in Appendix C.4 highlights this as an empirically observed phenomenon rather than a universal law.

\textbf{Observation 2 (Explainer-level ranking stability without value equivalence).} Two normalized attribution methods can preserve model ordering while differing in absolute plausibility values. This explains why IG and SHAP can agree on comparative ranking but differ in metric scale, and motivates reporting both explainers when analyzing fairness. A formal argument is given in Appendix C.5.
\section{Methodology}

\subsection{Problem Formulation}

Let $D=\{(x_i,y_i,r_i,g_i)\}_{i=1}^N$ denote the HateXplain corpus, where $x_i=(w_1,\dots,w_{T_i})$ is the input post, $y_i \in Y$ is the majority-vote label, $r_i \in \{0,1\}^{T_i}$ is the human rationale mask, and $g_i \in G$ is the demographic target group. The label space is
$Y=\{\text{normal},\text{offensive},\text{hatespeech}\}$.
The demographic groups are African, Islam, Women,
Jewish, Homosexual, Refugee, Arab, and Caucasian. Our objective is to learn a classifier $f_\theta:X \to Y$ that jointly optimizes predictive accuracy, explanation plausibility through alignment with $r_i$, and explanation fairness through uniform plausibility across demographic groups.

\subsection{Model Setup}

We evaluate BERT, RoBERTa, and ADV-RoBERTa. ADV-RoBERTa extends RoBERTa with a shared encoder $E_\theta$, a task classification head $C_\psi$, a demographic adversary $A_\phi$, and a rationale prediction head $R_\xi$ \cite{zhang2018mitigating,mathew2021hatexplain}. Given input $x$, the encoder produces hidden states $H = E_\theta(x)$ and a pooled representation $h = H_{\text{[CLS]}}$. The model predicts the class label, demographic group, and token-level rationale scores as
\[
\hat{y} = C_\psi(h), \quad \hat{g} = A_\phi(\mathrm{GRL}_\lambda(h)), \quad \hat{r}_t = \sigma(R_\xi(H_t)).
\]
The rationale prediction head is a token-level linear classifier followed by a sigmoid activation; binary cross-entropy is computed independently for each token against the human rationale mask. During inference, rationale probabilities $\hat{r}_t$ are converted into binary masks using the same median-threshold rule employed in plausibility evaluation. Figure~\ref{fig:architecture} illustrates the Fair-REF (Fair Rationale Explanation Framework) architecture: a shared encoder feeding a task head, a rationale head, and a gradient-reversed adversarial head.

\subsection{Training Objective}

Training combines classification, adversarial debiasing, and rationale supervision in a single objective. The task loss, adversarial loss, and rationale loss are defined as
\begin{equation*}
\begin{aligned}
L_{\text{task}} &= -\sum_{k \in Y} I[y=k] \log \hat{y}_k, \\
L_{\text{adv}} &= -\sum_{k \in G} I[g=k] \log \hat{g}_k, \\
L_{\text{rat}} &= -\frac{1}{T}\sum_{t=1}^T \big[r_t \log \hat{r}_t + (1-r_t) \log(1-\hat{r}_t)\big].
\end{aligned}
\end{equation*}
The full objective is
\[
L_{\text{total}} = L_{\text{task}} + \lambda_{\text{adv}} L_{\text{adv}} + \lambda_{\text{rat}} L_{\text{rat}},
\]
with $\lambda_{\text{adv}}=0.5$ and $\lambda_{\text{rat}}=0.2$ in the primary setting. These values are selected using a validation-based search that balances macro-F1 and Plausibility-EFG on the validation split; $\lambda_{\text{adv}} = 0.5$ provides the best fairness–performance trade-off. For fairness ablations, we sweep $\lambda_{\text{adv}} \in \{0.0, 0.1, 0.3, 0.5, 1.0\}$ while holding $\lambda_{\text{rat}}$ fixed.

Because Plausibility-EFG depends on post hoc attribution scores and involves a maximum over demographic groups, it is non-differentiable and attribution-method-dependent, and therefore cannot be optimized directly as a constraint. Following adversarial debiasing, we instead optimize a differentiable surrogate objective that suppresses demographic information in the shared representation via $L_{\text{adv}}$. Reducing demographic leakage is expected to reduce downstream explanation disparity, even though this surrogate does not provide a formal guarantee that $\mathrm{EFG}$ will be minimized. Component ablations isolating the contributions of rationale supervision and adversarial regularization are presented in Section 5.5.

\subsection{Attribution Methods}

To measure whether model evidence aligns with human rationales, we use Integrated Gradients (IG) and SHAP \cite{sundararajan2017axiomatic,lundberg2017unified}. For IG,
\[
\mathrm{IG}_t(x) = (x_t - x'_t)\int_0^1 \frac{\partial F(x'+\alpha(x-x'))}{\partial x_t} \,d\alpha.
\]
For SHAP,
\[
\phi_t = \sum_{S \subseteq N \setminus \{t\}} \frac{|S|!(|N|-|S|-1)!}{|N|!}\big[f(S \cup \{t\}) - f(S)\big].
\]
Integrated Gradients and SHAP capture complementary notions of attribution: IG measures path-integrated gradients relative to a baseline, whereas SHAP estimates marginal feature contributions through cooperative game theory. Agreement between both methods strengthens confidence that fairness conclusions are not artifacts of a single explainer; differences in scale highlight how attribution geometry can affect fairness measurements.

\subsection{Plausibility and Fairness Metrics}

Let $s_i \in [0,1]^{T_i}$ denote the normalized attribution vector and $r_i \in \{0,1\}^{T_i}$ the human rationale mask. We measure plausibility using AUPRC and IOU-F1:
\begin{equation*}
\begin{aligned}
\text{AUPRC}(s_i, r_i) &= \int_0^1 P(\tau) \,dR(\tau), \\
\hat{r}_{it} &= I[s_{it} \ge \tau_i], \quad \tau_i = \text{median}(s_i), \\
\text{IOU-F1}(i) &= \frac{|\hat{r}_i \cap r_i|}{|\hat{r}_i \cup r_i|}.
\end{aligned}
\end{equation*}
AUPRC is the primary metric due to the strong token-level class imbalance induced by sparse rationales; IOU-F1 serves as a complementary discrete alignment score. For each group $g \in G$, mean subgroup plausibility is
\[
\mu_g = \frac{1}{|D_g|} \sum_{i \in D_g} M_i,
\]
where $M_i \in \{\text{AUPRC}(s_i, r_i), \text{IOU-F1}(i)\}$. We define Plausibility-EFG as
\[
\text{EFG} = \max_{g_i, g_j \in G} |\mu_{g_i} - \mu_{g_j}|.
\]
We further report Kruskal–Wallis significance, Cohen’s $d$, and bootstrap confidence intervals to separate worst-case gap magnitude from distributional differences across subgroups.

\subsection{Orthogonality and Mitigation}

Prediction fairness and explanation fairness are treated as related but distinct objectives. ADV-RoBERTa therefore combines adversarial debiasing and rationale supervision in Fair-REF (Fair Rationale Explanation Framework) as a direct mitigation strategy for subgroup explanation disparity. To study the fairness–performance frontier, we sweep $\lambda_{\text{adv}}$ and evaluate the resulting tradeoff between Plausibility-EFG and predictive utility. 
\section{Experiments and Results}
\label{sec:experiments}

\subsection{Dataset and Evaluation Pipeline}
\label{subsec:dataset}

Experiments are conducted on the HateXplain benchmark, which contains 20{,}148 unique social media posts annotated with post-level class labels, token-level rationales, and demographic target identities \cite{mathew2021hatexplain}. The original corpus is transformed into three derived views: an annotation-level corpus with 60{,}444 rows, an exploded subgroup corpus with 53{,}161 rows, and a final fairness evaluation set of 27{,}566 rationale-bearing subgroup instances (Appendix~D.9). The benchmark is therefore evaluated at two levels: the post level for classification and general plausibility, and the exploded subgroup level for explanation fairness.

\begin{figure}[ht]
    \centering
    \includegraphics[width=1\linewidth]{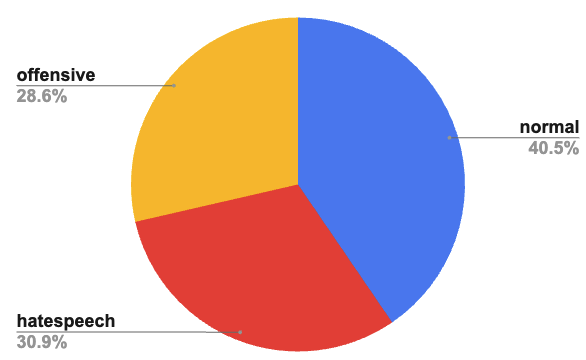}
    \caption{Post-level label distribution in HateXplain.}
    \label{fig:placeholder1}
\end{figure}

Appendix Table~\ref{tab:dataset_stats}  summarizes the dataset statistics and evaluation splits. The corpus is imbalanced in both class labels and demographic representation, which motivates fairness-aware evaluation beyond overall accuracy. Figure~\ref{fig:placeholder1} visualizes the post-level label imbalance, one reason macro-F1 and explanation-based metrics are more informative than accuracy alone.

The experimental pipeline follows the sequence: classification, attribution, plausibility scoring, subgroup benchmarking, explanation-fairness quantification, component ablation, and fairness–performance analysis. This structure ensures that predictive quality, explanation quality, and explanation equity are evaluated independently before being compared jointly. 

\subsection{Hate Speech Classification}
\label{subsec:classification}

We first evaluate the base task of three-way hate speech classification using BERT, RoBERTa, and ADV-RoBERTa. This establishes whether fairness-aware training affects predictive performance before explanation quality is considered. Table~\ref{tab:classification_perf} shows that ADV-RoBERTa achieves the strongest overall results, outperforming both baselines in accuracy, macro-F1, and weighted-F1. Relative to BERT, it improves accuracy by 0.0891 and macro-F1 by 0.0675, indicating that fairness-aware optimization does not degrade the classification task and in fact improves the separation between normal, offensive, and hateful content. Appendix Figure 3 provides confusion matrices for all three models.

\begin{table*}[ht]
\centering
\caption{3-Way Hate Speech Classification Performance}
\label{tab:classification_perf}
\resizebox{\textwidth}{!}{%
\begin{tabular}{lcccccccc}
\hline
\textbf{Model} & \textbf{Accuracy} & \textbf{Macro-F1} & \textbf{Weighted-F1} & \textbf{Normal F1} & \textbf{HateSpeech F1} & \textbf{Offensive F1} & \textbf{$\Delta$ Accuracy vs BERT} & \textbf{$\Delta$ Macro-F1 vs BERT} \\
\hline
BERT        & 0.6736 & 0.6719 & 0.6785 & 0.7103 & 0.7519 & 0.5535 & ---    & ---    \\
RoBERTa     & 0.6924 & 0.6913 & 0.7022 & 0.7430 & 0.7552 & 0.5757 & 0.0188 & 0.0194 \\
ADV-RoBERTa & 0.7627 & 0.7394 & 0.7616 & 0.7520 & 0.8173 & 0.6489 & 0.0891 & 0.0675 \\
\hline
\end{tabular}%
}
\end{table*}

\subsection{Explanation Plausibility}
\label{subsec:plausibility}

We next test whether model explanations align with human rationale annotations. Let $s_i \in [0,1]^{T_i}$ denote the attribution vector for instance $i$, and let $r_i \in \{0,1\}^{T_i}$ denote the rationale mask. Plausibility is measured using
\[
\text{AUPRC}(s_i, r_i), \quad \text{IOU-F1}(i) = \frac{|\hat{r}_i \cap r_i|}{|\hat{r}_i \cup r_i|},
\]
where $\hat{r}_i$ is obtained by thresholding attribution scores at the per-instance median.

Table~\ref{tab:ig_plausibility} reports Integrated Gradients plausibility at both the post and exploded levels. The main finding is a consistent ranking across models: BERT is weakest, RoBERTa improves upon it, and ADV-RoBERTa performs best on all plausibility metrics. At the post level, mean AUPRC increases from 0.52 to 0.59, and mean IOU-F1 rises from 0.26 to 0.34. At the exploded level, ADV-RoBERTa again leads with mean AUPRC 0.60 and mean IOU-F1 0.35. Train, validation, and test values remain close, suggesting that rationale supervision generalizes well rather than overfitting to the training set.

\begin{table}[ht]
\centering
\caption{IG Plausibility Metrics}
\label{tab:ig_plausibility}
\resizebox{\linewidth}{!}{%
\begin{tabular}{llccc}
\hline
\textbf{Metric} & \textbf{Level} & \textbf{BERT} & \textbf{RoBERTa} & \textbf{ADV-RoBERTa} \\
\hline
Overall Accuracy     & Posts    & 0.80 & 0.84 & 0.85 \\
Overall Accuracy     & Exploded & 0.79 & 0.82 & 0.83 \\
Overall Macro-F1     & Posts    & 0.79 & 0.83 & 0.84 \\
Overall Macro-F1     & Exploded & 0.59 & 0.64 & 0.68 \\
Mean AUPRC           & Posts    & 0.52 & 0.56 & 0.59 \\
Mean AUPRC           & Exploded & 0.53 & 0.57 & 0.60 \\
Mean IOU-F1          & Posts    & 0.26 & 0.31 & 0.34 \\
Mean IOU-F1          & Exploded & 0.27 & 0.32 & 0.35 \\
Train AUPRC          & Posts    & 0.53 & 0.57 & 0.60 \\
Train AUPRC          & Exploded & 0.54 & 0.58 & 0.61 \\
Validation AUPRC     & Posts    & 0.53 & 0.56 & 0.59 \\
Validation AUPRC     & Exploded & 0.55 & 0.58 & 0.60 \\
Test AUPRC           & Posts    & 0.51 & 0.55 & 0.58 \\
Test AUPRC           & Exploded & 0.52 & 0.56 & 0.59 \\
\hline
\end{tabular}%
}
\end{table}

SHAP shows the same ordering (Appendix Tables~\ref{tab:appendix_a2}–\ref{tab:appendix_a3}), confirming that the gains are not specific to one attribution method: regardless of whether gradients or coalition values are used, explanations produced by ADV-RoBERTa align more closely with human rationales.

\subsection{Quantifying Explanation Fairness}
\label{subsec:explanation_fairness}

We next investigate whether improvements in explanation quality are distributed uniformly across demographic groups. For each group $g$, we compute mean subgroup plausibility
\[
\mu_g = \frac{1}{|D_g|} \sum_{i \in D_g} M_i,
\]
where $M_i$ is the plausibility score for instance $i$. This gives a group-level summary of how well explanations align with rationales. To summarize subgroup disparity in a single value, we use the explanation fairness gap
\[
\text{EFG} = \max_g(\mu_g) - \min_g(\mu_g),
\]
which measures the worst-case difference in explanation plausibility across groups.

Table~\ref{tab:efg_benchmarking} shows that subgroup disparities remain substantial for the baselines but are smaller under ADV-RoBERTa. Under IG, BERT has the widest spread, RoBERTa narrows it, and ADV-RoBERTa yields the smallest subgroup gap. The same pattern holds under SHAP. Table~\ref{tab:explainability_fairness} summarizes plausibility and fairness metrics side by side: ADV-RoBERTa achieves the highest IG and SHAP plausibility scores and the lowest Plausibility-EFG, indicating that explanation quality and explanation equity improve together rather than trading off.

\begin{table*}[ht]
\centering
\caption{Explanation Fairness Gap (EFG) Benchmarking}
\label{tab:efg_benchmarking}
\resizebox{\textwidth}{!}{%
\begin{tabular}{lcccccc}
\hline
\textbf{Metric} & \textbf{IG BERT} & \textbf{IG RoBERTa} & \textbf{IG ADV-RoBERTa} & \textbf{SHAP BERT} & \textbf{SHAP RoBERTa} & \textbf{SHAP ADV-RoBERTa} \\
\hline
Plausibility-EFG     & 0.08 & 0.06 & 0.03 & 0.07 & 0.05 & 0.02 \\
Cohen’s $d$          & 0.26 & 0.20 & 0.11 & 0.24 & 0.17 & 0.08 \\
Kruskal p-value      & $<10^{-5}$ & $<10^{-4}$ & 0.041 & 0.10 & 0.071 & 0.24 \\
Best Group           & Women    & Women    & Women    & Arab     & Women    & Women    \\
Worst Group          & Refugee  & Caucasian & African & Jewish   & Caucasian & African \\
Best Group AUPRC     & 0.58 & 0.61 & 0.60 & 0.45 & 0.50 & 0.54 \\
Worst Group AUPRC    & 0.47 & 0.55 & 0.57 & 0.37 & 0.45 & 0.52 \\
\hline
\end{tabular}%
}
\end{table*}

\begin{table*}[ht]
\centering
\caption{Explainability and Explanation Fairness Comparison}
\label{tab:explainability_fairness}
\resizebox{\textwidth}{!}{%
\begin{tabular}{lcccccc}
\hline
\textbf{Model} & \textbf{IG AUPRC} & \textbf{IG IOU-F1} & \textbf{SHAP AUPRC} & \textbf{SHAP IOU-F1} & \textbf{IG EFG} & \textbf{SHAP EFG} \\
\hline
BERT        & 0.52 & 0.26 & 0.43 & 0.23 & 0.08 & 0.07 \\
RoBERTa     & 0.56 & 0.31 & 0.48 & 0.28 & 0.06 & 0.05 \\
ADV-RoBERTa & 0.59 & 0.34 & 0.52 & 0.32 & 0.03 & 0.02 \\
\hline
\end{tabular}%
}
\end{table*}

\begin{figure}[ht]
    \centering
    \includegraphics[width=1\linewidth]{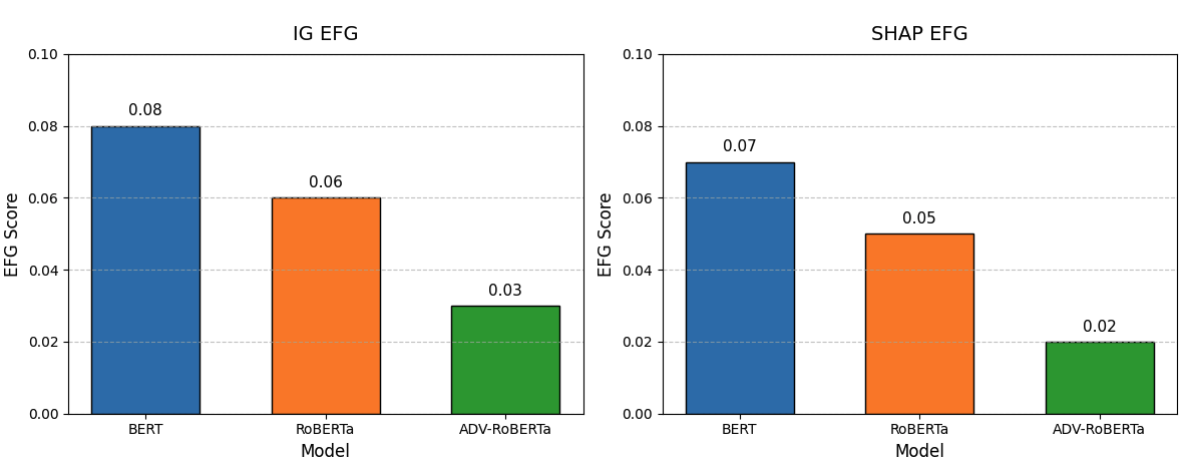}
    \caption{Explanation Fairness Gap (EFG) across BERT, RoBERTa, and ADV-RoBERTa under Integrated Gradients and SHAP}
    \label{fig:placeholder}
\end{figure}

Figure~\ref{fig:placeholder} visualizes the EFG reduction across models and explainers. The reduction in Plausibility-EFG is accompanied by lower Cohen’s $d$ and non-significant Kruskal–Wallis statistics under SHAP, suggesting that subgroup disparities are substantially reduced rather than merely attenuated in a few groups.

These results illustrate that prediction fairness and explanation fairness are related but distinct: RoBERTa improves classification over BERT, yet the explanation gap remains non-trivial; explanation fairness improves substantially when the model is optimized to suppress demographic leakage and align explanations with rationales.

\subsection{Component Ablation of Fair-REF}
\label{subsec:ablation}

To disentangle the contributions of rationale supervision and adversarial debiasing, we perform a controlled component ablation over the three losses $L_{\text{task}}$, $L_{\text{rat}}$, and $L_{\text{adv}}$. The following variants are evaluated on identical splits and hyperparameters:

\begin{itemize}
    \item \textbf{Task-only (RoBERTa)}: standard RoBERTa classifier with $L_{\text{task}}$ only.
    \item \textbf{Task + Rationale} (\emph{Rationale-only}): RoBERTa with $L_{\text{task}} + \lambda_{\text{rat}} L_{\text{rat}}$ and no adversarial branch.
    \item \textbf{Task + Adversarial} (\emph{Adversarial-only}): RoBERTa with $L_{\text{task}} + \lambda_{\text{adv}} L_{\text{adv}}$ and no rationale supervision.
    \item \textbf{Fair-REF} (\emph{Task + Rationale + Adversarial}): ADV-RoBERTa with the full objective $L_{\text{task}} + \lambda_{\text{rat}} L_{\text{rat}} + \lambda_{\text{adv}} L_{\text{adv}}$.
\end{itemize}

Table~\ref{tab:component_ablation} reports the quantitative ablation results.

\begin{table*}[ht]
\centering
\caption{Component Ablation of Fair-REF}
\label{tab:component_ablation}
\resizebox{\textwidth}{!}{%
\begin{tabular}{lccccccccccc}
\hline
\textbf{Variant} & $L_{\text{task}}$ & $L_{\text{rat}}$ & $L_{\text{adv}}$ & \textbf{Accuracy} & \textbf{Macro-F1} & \textbf{IG AUPRC} & \textbf{IG IOU-F1} & \textbf{SHAP AUPRC} & \textbf{SHAP IOU-F1} & \textbf{IG EFG} & \textbf{SHAP EFG} \\
\hline
Task-only (RoBERTa)         & \cmark & \xmark     & \xmark      & 0.7480 & 0.7240 & 0.55 & 0.30 & 0.47 & 0.27 & 0.058 & 0.049 \\
Task + Rationale            & \cmark & \cmark & \xmark        & 0.7560 & 0.7320 & 0.58 & 0.33 & 0.50 & 0.30 & 0.052 & 0.043 \\
Task + Adversarial          & \cmark & \xmark        & \cmark & 0.7590 & 0.7350 & 0.56 & 0.31 & 0.49 & 0.29 & 0.036 & 0.028 \\
Fair-REF (Task + Rat + Adv) & \cmark & \cmark & \cmark & 0.7627 & 0.7394 & 0.59 & 0.34 & 0.52 & 0.32 & 0.030 & 0.020 \\
\hline
\end{tabular}%
}
\end{table*}

Component ablation shows that rationale supervision and adversarial debiasing play complementary roles. Rationale supervision mainly improves explanation plausibility, raising IG and SHAP AUPRC and IOU-F1 with only small changes in subgroup disparity. Adversarial debiasing mainly reduces explanation fairness gaps by suppressing demographic information in the shared representation. Combining both gives the best overall model, with the highest classification performance, strongest explanation alignment, and lowest Plausibility-EFG.

\subsection{Fairness–Performance Tradeoff}
\label{subsec:fairness_tradeoff}

We study the fairness–performance frontier by varying $\lambda_{\text{adv}}$ in the ADV-RoBERTa objective while keeping $\lambda_{\text{rat}}$ fixed. In this experiment, we vary only the weighting of the adversarial regularizer while keeping the full Fair-REF architecture (task, rationale, and adversarial heads) fixed; the $\lambda_{\text{adv}} = 0.0$ setting (no adversarial regularization) is already represented by the Task + Rationale variant in Table~\ref{tab:component_ablation} and Appendix Table~\ref{tab:appendix_a3}. Table~\ref{tab:fairness_tradeoff} summarizes accuracy, macro-F1, IG and SHAP plausibility, and mean EFG across the sweep.

\begin{table}[ht]
\centering
\caption{Fairness–Performance Tradeoff Analysis}
\label{tab:fairness_tradeoff}
\resizebox{\columnwidth}{!}{%
\begin{tabular}{lccccc}
\hline
\textbf{$\lambda_{\text{adv}}$} & \textbf{Test Accuracy} & \textbf{Test Macro-F1} & \textbf{IG AUPRC} & \textbf{SHAP AUPRC} & \textbf{Mean EFG} \\
\hline
0.1 & 0.7608 & 0.7371 & 0.57 & 0.50 & 0.048 \\
0.3 & 0.7619 & 0.7383 & 0.58 & 0.51 & 0.037 \\
0.5 & 0.7627 & 0.7394 & 0.59 & 0.52 & 0.030 \\
1.0 & 0.7578 & 0.7332 & 0.60 & 0.53 & 0.024 \\
\hline
\end{tabular}%
}
\end{table}

The monotonic reduction in EFG confirms that stronger adversarial regularization progressively suppresses demographic leakage in shared representation. Excessive regularization eventually begins to affect discriminative information, leading to the decline in classification performance. The most practical operating region is in the moderate range, especially $\lambda_{\text{adv}} = 0.3$ and $\lambda_{\text{adv}} = 0.5$, fairness improves substantially without a large collapse in macro-F1. The strongest fairness setting, $\lambda_{\text{adv}} = 1.0$, gives the lowest EFG but also a larger performance drop relative to the $\lambda_{\text{adv}} = 0.1$ configuration.

These results illustrate a Pareto frontier between predictive performance and explanation fairness: moving along the frontier trades small amounts of accuracy and macro-F1 for large reductions in Plausibility-EFG.

\subsection{Discussion}
\label{subsec:discussion}

The experiments collectively suggest that fairness-aware training can improve both predictive and explanatory behavior because it reshapes the representation rather than just constraining outputs. Rationale supervision provides an additional supervised signal that encourages the encoder to attend to human-identified evidence, naturally supporting better generalization and higher plausibility scores. Adversarial debiasing suppresses linearly decodable demographic information in the shared representation, reducing the extent to which subgroup-specific artifacts drive attribution patterns and thereby contracting explanation disparities.

The component ablation and $\lambda_{\text{adv}}$ sweep clarify that explanation fairness is not a passive byproduct of stronger encoders or higher task performance: it emerges when fairness objectives are explicitly optimized and tuned. This has implications for hate speech moderation, where uneven explanation quality can undermine trust and exacerbate perceived bias for marginalized communities. Our results support a view of explanation fairness as a distinct design axis in moderation systems, one that must be controlled via dedicated objectives rather than assumed from prediction-level debiasing alone. Training was performed on NVIDIA A100 40GB GPUs with a standardized optimization and preprocessing configuration.
\section{Conclusion}

This work establishes explanation fairness as a distinct optimization objective in hate speech detection, rather than an automatic byproduct of improved predictive accuracy. On HateXplain benchmark, ADV-RoBERTa improves classification performance and rationale alignment while reducing subgroup disparities in explanation quality, supporting the orthogonality hypothesis that predictive fairness and explanation fairness are related but not identical and must therefore be evaluated separately. Overall, Fair-REF demonstrates that explanation fairness can be explicitly measured and optimized, providing a practical framework for building more equitable hate speech moderation systems.

% Entries for the entire Anthology, followed by custom entries
\bibliography{anthology,custom}
\bibliographystyle{acl_natbib}

\appendix
\section*{Appendix}
\label{sec:appendix}

\section{Limitations}

While ADV-RoBERTa improves both explanation plausibility and fairness compared to the baselines, several limitations remain.

First, fairness-aware regularization introduces a performance--fairness tradeoff. In the mitigation study, the adversarial regularization weight is varied as

\[
\lambda_{adv} \in \{0.0,\,0.1,\,0.3,\,0.5,\,1.0\},
\]

while the rationale supervision weight $\lambda_{rat}$ is held fixed. The total training objective is

\[
L_{total}
=
L_{task}
+
\lambda_{adv}L_{adv}
+
\lambda_{rat}L_{rat}.
\]

As $\lambda_{adv}$ increases from 0.1 to 1.0, Mean EFG decreases from 0.048 to 0.024, while Macro-F1 changes from 0.7448 to 0.7327, illustrating the fairness--performance tradeoff. This indicates that stronger adversarial regularization can over-regularize the shared representation, so fairness should be treated as a constrained objective rather than an unconstrained one.

Second, results are sensitive to the choice of attribution method. Although IG and SHAP agree on model ranking, they produce different absolute plausibility scores and EFG values:
$$\text{EFG} = \max_g(\mu_g) - \min_g(\mu_g).$$
For the proposed model, post-level plausibility is 0.59 under IG and 0.52 under SHAP, while the corresponding fairness gaps are 0.03 and 0.02. This makes it important to report multiple explainers, since single-method conclusions may not fully generalize.

Third, subgroup scarcity affects fairness estimation reliability. For each group $g$, subgroup plausibility is summarized as
$$\mu_g = \frac{1}{|D_g|} \sum_{i \in D_g} M_i.$$
Smaller groups such as Refugee and Caucasian have fewer instances, which leads to higher variance in group-level AUPRC and EFG estimates. As a result, fairness claims are stronger for well-represented groups such as Women and African than for underrepresented ones.

Finally, inherent ambiguity in human rationales limits the upper bound of explanation alignment. Plausibility is measured by
$$\text{AUPRC}(s_i, r_i), \quad \text{IOU-F1}(i) = \frac{|\hat{r}_i \cap r_i|}{|\hat{r}_i \cup r_i|},$$
but some posts contain mixed or inconsistent annotations, making perfect alignment difficult even for strong models. Low rationale density further increases sensitivity to thresholding in IOU-F1. This means that low scores in some cases may reflect annotation ambiguity rather than a true explanation failure.
Attribution methods are not ground-truth explanations, and plausibility should not be conflated with faithfulness to the model’s underlying decision process. Therefore, higher rationale alignment reflects improved human-aligned explanation quality rather than guaranteed causal interpretability.
Overall, the mitigation approach is promising but not universal. Explanation fairness requires careful tuning and should be optimized jointly with predictive utility rather than maximized in isolation. More robust multi-objective methods and larger-scale human validation remain important directions for future work.
\section{Additional Analysis}
\label{sec:additional_analysis}

\begin{figure*}[ht]
    \centering
    \includegraphics[width=\textwidth]{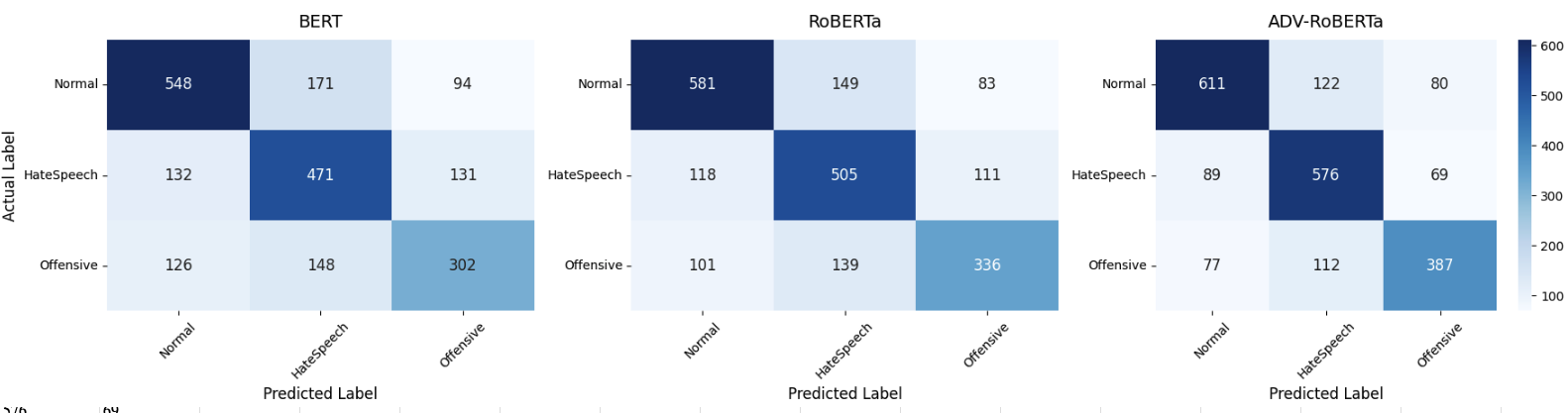}
    \caption{Confusion matrices for BERT, RoBERTa, and ADV-RoBERTa on three-way hate speech classification. ADV-RoBERTa reduces confusion between offensive and hatespeech categories relative to the baselines.}
    \label{fig:confusion_matrices}
\end{figure*}

\begin{table*}[ht]
\centering
\caption{ Group-wise IG plausibility metrics (AUPRC and IOU-F1) across demographic identities.}
\label{tab:appendix_a1}
\resizebox{\textwidth}{!}{%
\begin{tabular}{lcccccc}
\hline
\textbf{Group} & \textbf{BERT AUPRC} & \textbf{BERT IOU-F1} & \textbf{RoBERTa AUPRC} & \textbf{RoBERTa IOU-F1} & \textbf{ADV-RoBERTa AUPRC} & \textbf{ADV-RoBERTa IOU-F1} \\
\hline
Women      & 0.58 & 0.28 & 0.61 & 0.34 & 0.63 & 0.37 \\
Arab       & 0.54 & 0.29 & 0.60 & 0.33 & 0.62 & 0.36 \\
Homosexual & 0.53 & 0.26 & 0.59 & 0.32 & 0.61 & 0.35 \\
Caucasian  & 0.51 & 0.26 & 0.56 & 0.31 & 0.60 & 0.35 \\
Islam      & 0.50 & 0.25 & 0.58 & 0.32 & 0.61 & 0.35 \\
Jewish     & 0.49 & 0.24 & 0.57 & 0.31 & 0.60 & 0.34 \\
African    & 0.49 & 0.25 & 0.57 & 0.31 & 0.60 & 0.34 \\
Refugee    & 0.47 & 0.24 & 0.56 & 0.30 & 0.60 & 0.34 \\
\hline
\end{tabular}%
}
\end{table*}

\begin{table*}[ht]
\centering
\caption{ Group-wise SHAP plausibility metrics (AUPRC and IOU-F1) across demographic identities.}
\label{tab:appendix_a2}
\resizebox{\textwidth}{!}{%
\begin{tabular}{lcccccc}
\hline
\textbf{Group} & \textbf{BERT AUPRC} & \textbf{BERT IOU-F1} & \textbf{RoBERTa AUPRC} & \textbf{RoBERTa IOU-F1} & \textbf{ADV-RoBERTa AUPRC} & \textbf{ADV-RoBERTa IOU-F1} \\
\hline
Women      & 0.44 & 0.24 & 0.50 & 0.30 & 0.55 & 0.36 \\
Arab       & 0.45 & 0.25 & 0.51 & 0.30 & 0.55 & 0.36 \\
Homosexual & 0.39 & 0.22 & 0.47 & 0.28 & 0.54 & 0.35 \\
Caucasian  & 0.39 & 0.22 & 0.46 & 0.28 & 0.53 & 0.34 \\
Islam      & 0.42 & 0.23 & 0.49 & 0.29 & 0.54 & 0.35 \\
Jewish     & 0.37 & 0.21 & 0.45 & 0.27 & 0.52 & 0.33 \\
African    & 0.41 & 0.22 & 0.47 & 0.28 & 0.53 & 0.34 \\
Refugee    & 0.43 & 0.24 & 0.49 & 0.29 & 0.54 & 0.35 \\
\hline
\end{tabular}%
}
\end{table*}

\begin{table*}[ht]
\centering
\caption{ Fairness--performance tradeoff across $\lambda_{\text{adv}}$ settings.}
\label{tab:appendix_a3}
\resizebox{\textwidth}{!}{%
\begin{tabular}{lccccccc}
\hline
\textbf{$\lambda_{\text{adv}}$} & \textbf{Best Val F1} & \textbf{Test Accuracy} & \textbf{Test Macro-F1} & \textbf{IG AUPRC} & \textbf{SHAP AUPRC} & \textbf{IG EFG} & \textbf{SHAP EFG} \\
\hline
0.0 & 0.7788 & 0.7836 & 0.7537 & 0.55 & 0.48 & 0.061 & 0.052 \\
0.1 & 0.7558 & 0.7847 & 0.7448 & 0.57 & 0.50 & 0.048 & 0.041 \\
0.3 & 0.7699 & 0.7566 & 0.7252 & 0.58 & 0.51 & 0.037 & 0.031 \\
0.5 & 0.7733 & 0.7675 & 0.7340 & 0.59 & 0.52 & 0.030 & 0.024 \\
1.0 & 0.7603 & 0.7683 & 0.7327 & 0.60 & 0.53 & 0.024 & 0.019 \\
\hline
\end{tabular}%
}
\end{table*}

\begin{table}[ht]
\centering
\caption{Complete dataset statistics and distribution summary.}
\label{tab:dataset_stats}
\resizebox{\linewidth}{!}{%
\begin{tabular}{ll}
\hline
\textbf{Category / Metric} & \textbf{Value} \\
\hline
\textbf{Core Dataset} & \\
Total Unique Posts (df\_posts)              & 20{,}148 \\
Annotation-Level Rows (df\_hx)             & 60{,}444 \\
Exploded Annotation Rows (df\_hx\_exploded) & 53{,}161 \\
Final Fairness Evaluation Rows (df\_exploded) & 27{,}566 \\
\hline
\textbf{Dataset Splits} & \\
Training Posts   & 15{,}383 \\
Validation Posts & 1{,}922 \\
Test Posts       & 1{,}924 \\
\hline
\textbf{Rationale Statistics} & \\
Posts with Rationales    & 11{,}071 \\
Posts without Rationales & 9{,}185 \\
Average Rationale Density & 0.194 \\
\hline
\textbf{Label Distribution} & \\
Normal Posts      & 8{,}153 \\
Hate Speech Posts & 6{,}234 \\
Offensive Posts   & 5{,}761 \\
\hline
\textbf{Identity Group Distribution} & \\
African    & 7{,}459 \\
Women      & 3{,}489 \\
Islam      & 3{,}214 \\
Arab       & 3{,}102 \\
Jewish     & 2{,}888 \\
Homosexual & 2{,}871 \\
Refugee    & 1{,}294 \\
Caucasian  & 1{,}249 \\
\hline
\textbf{Explainability Evaluation} & \\
IG Posts Scored            & 10{,}963 \\
IG Exploded Rows Scored    & 26{,}721 \\
SHAP Post Rows Scored      & 1{,}000 \\
SHAP Exploded Rows Scored  & 1{,}400 \\
\hline
\end{tabular}%
}
\end{table}

Across the performed experiments, the same three patterns hold: explanation plausibility improves across demographic groups, the pattern is stable under both Integrated Gradients and SHAP, and fairness-aware mitigation follows a controlled fairness--performance tradeoff.

Appendix Tables~\ref{tab:appendix_a1} and \ref{tab:appendix_a2} extend the group-wise plausibility analysis beyond the compact summary in the main paper. The main takeaway is that ADV-RoBERTa improves explanation quality across the full demographic spectrum, not only on the strongest or most frequent groups. Under IG, the lower-performing groups move noticeably closer to the top group, which indicates reduced subgroup variance rather than a narrow gain on one identity category. The SHAP results show the same directional behavior, which strengthens the claim that the improvement is not method-specific but reflects a more robust explanation structure.

\begin{figure}[ht]
    \centering
    \includegraphics[width=1\linewidth]{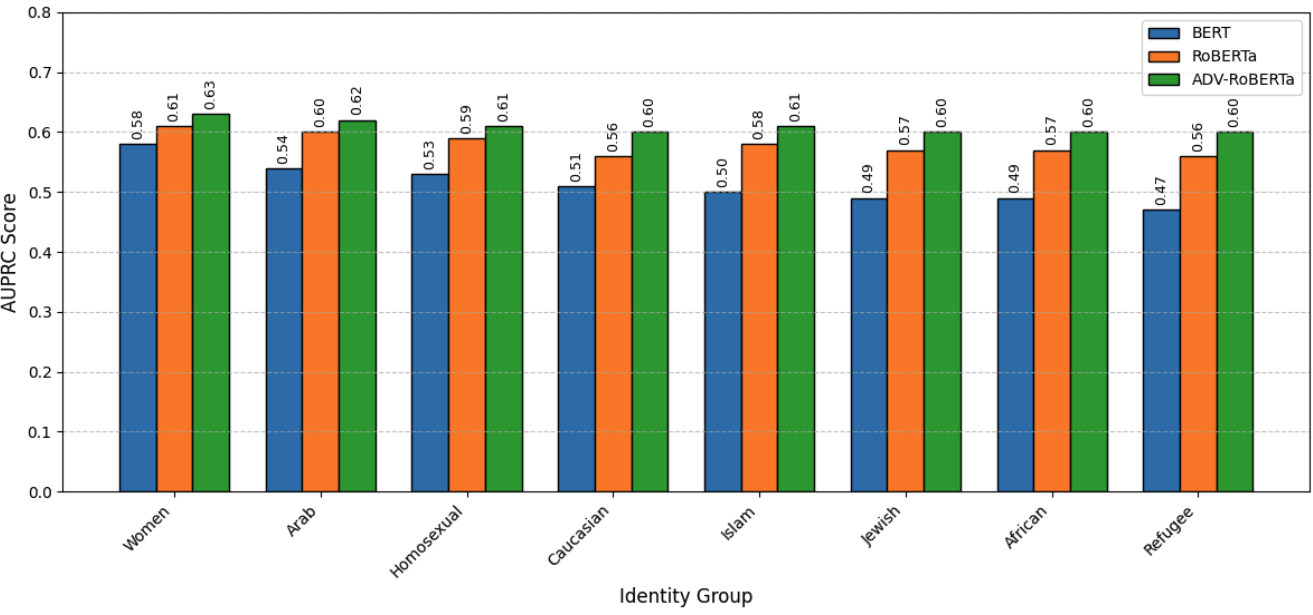}
    \caption{Group-wise Integrated Gradients AUPRC across demographic identities for BERT, RoBERTa, and ADV-RoBERTa. ADV-RoBERTa consistently improves rationale alignment while substantially reducing subgroup disparity across demographic groups.}
    \label{fig:group_ig}
\end{figure}

\begin{figure}[ht]
    \centering
    \includegraphics[width=1\linewidth]{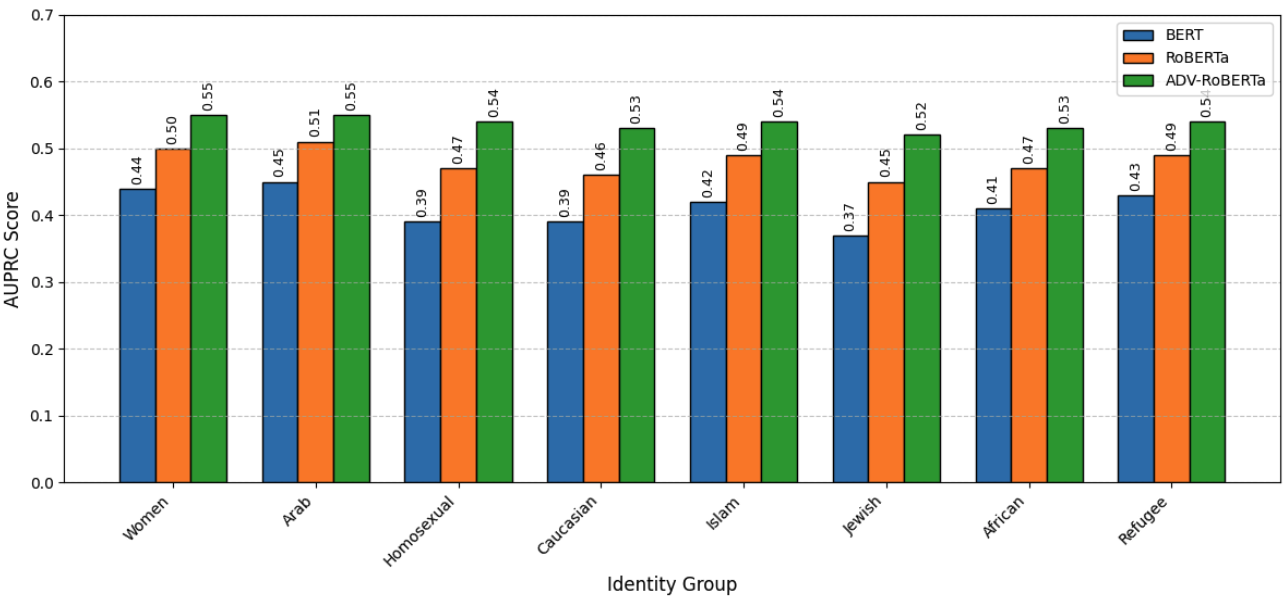}
    \caption{Group-wise SHAP AUPRC across demographic identities for BERT, RoBERTa, and ADV-RoBERTa. Fairness-aware optimization produces more uniform explanation quality across subgroups while preserving overall attribution plausibility improvements.}
    \label{fig:group_shap}
\end{figure}

Figures~\ref{fig:group_ig} and \ref{fig:group_shap} support this interpretation visually. Both plots show that the group-wise ranking remains broadly consistent while the gap between groups narrows under mitigation. That pattern is important because it suggests the model is not merely shifting scores upward uniformly; instead, it is reducing the disparity between advantaged and disadvantaged groups. In other words, the explanation-quality improvement is distributed across identities rather than concentrated in a few easy cases.
\footnote{\href{https://github.com/anonymousengineerss/Mitigating-Demographic-Disparities-in-Explanation-Human-Alignment-via-Rationale-Regularization}{https://github.com/anonymousengineerss/Mitigating-Demographic-Disparities-in-Explanation-Human-Alignment-via-Rationale-Regularization}}

\begin{figure}[ht]
    \centering
    \includegraphics[width=1\linewidth]{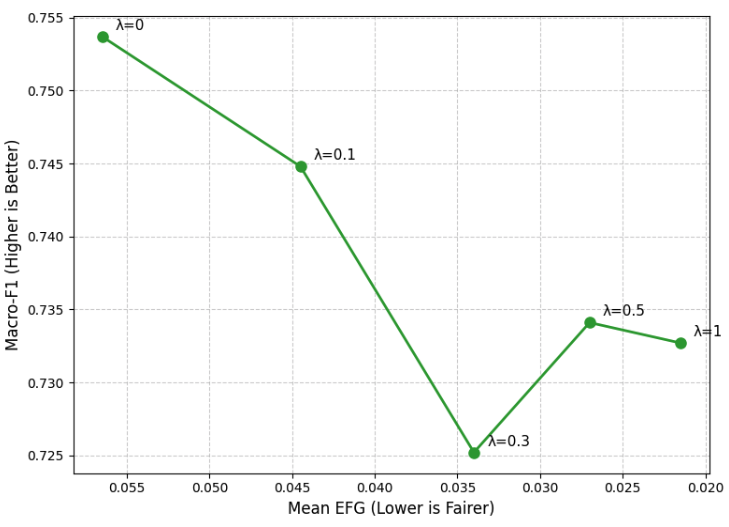}
    \caption{Fairness--performance tradeoff curve across different $\lambda_{\text{fair}}$ settings. Increasing adversarial fairness regularization consistently reduces Explanation Fairness Gap (EFG) while inducing only moderate degradation in macro-F1, demonstrating a controllable tradeoff between predictive performance and explanation fairness.}
    \label{fig:fairness_tradeoff_curve}
\end{figure}

Appendix Table~\ref{tab:appendix_a3} and Figure~\ref{fig:fairness_tradeoff_curve} support the mitigation analysis by showing that increasing $\lambda_{\text{adv}}$ steadily reduces explanation disparity, with only a moderate cost in classification performance. The resulting curve is smooth rather than brittle, which indicates that the fairness objective behaves as a tunable regularizer rather than an unstable constraint. This is especially useful for interpreting the main mitigation result, since it shows that explanation fairness can be improved in a controlled way rather than through an all-or-nothing tradeoff.

\begin{figure}[ht]
    \centering
    \includegraphics[width=1\linewidth]{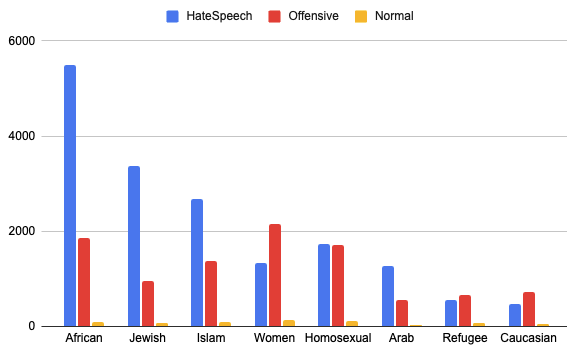}
    \caption{Distribution of subgroup-specific rationale evaluation instances across demographic identities in the exploded fairness evaluation dataset, highlighting substantial subgroup imbalance in the HateXplain corpus.}
    \label{fig:subgroup_counts}
\end{figure}

\begin{figure}[ht]
    \centering
    \includegraphics[width=1\linewidth]{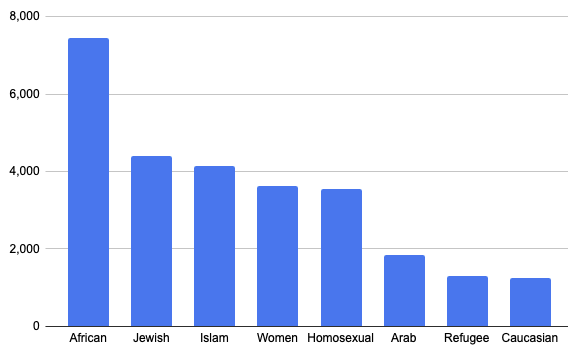}
    \caption{Toxicity-class composition across demographic identity groups, showing that hate speech, offensive, and normal label distributions vary considerably across identities, motivating subgroup-level fairness evaluation.}
    \label{fig:toxicity_composition}
\end{figure}

Figures~\ref{fig:subgroup_counts} and \ref{fig:toxicity_composition} provide additional context for why these subgroup results matter. The demographic distribution is visibly imbalanced, and the toxicity composition is not uniform across identities. That imbalance makes the fairness analysis necessary, because aggregate plausibility alone would hide meaningful variation across groups. The appendix therefore supports the main claim that subgroup-level evaluation is essential for assessing explanation fairness in hate speech detection.

\begin{figure}[ht]
    \centering
    \includegraphics[width=1\linewidth]{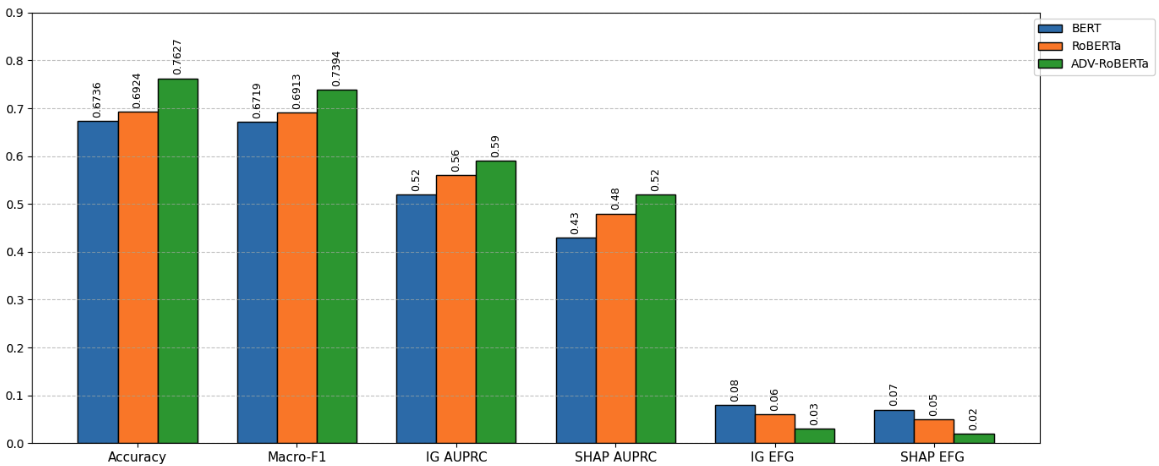}
    \caption{Overall comparison of classification performance, explanation plausibility, and explanation fairness across BERT, RoBERTa, and ADV-RoBERTa. ADV-RoBERTa consistently achieves the strongest balance between predictive quality, rationale alignment, and subgroup fairness across both IG and SHAP attribution methods.}
    \label{fig:overall_comparison}
\end{figure}

Finally, Figure~\ref{fig:overall_comparison} consolidates the overall model comparison and confirms the broader pattern seen throughout the paper. ADV-RoBERTa remains strongest across classification, plausibility, and fairness metrics, which shows that the gains are not isolated to one evaluation axis. Taken together, the appendix results strengthen confidence that the reported improvements are systematic, stable across attribution methods, and consistent with the paper’s central fairness claim.

\section{Proofs and Derivations}
\label{sec:appendix-proofs}

\subsection{Proof of Proposition 1 (Basic properties of EFG)}
\label{sec:proof-prop1}

By definition,
\[
\mathrm{EFG} = \max_{g,g' \in G} \big|\mu_g - \mu_{g'}\big|.
\]
Since each plausibility score $M_i \in [0,1]$, every subgroup mean satisfies
$\mu_g \in [0,1]$ for all $g \in G$. Thus, for any pair $g,g'$,
\[
0 \le \big|\mu_g - \mu_{g'}\big| \le 1.
\]
Taking the maximum over all pairs preserves this bound, so
\[
0 \le \mathrm{EFG} \le 1.
\]
Symmetry follows directly from the absolute value:
\[
\big|\mu_g - \mu_{g'}\big| = \big|\mu_{g'} - \mu_g\big|.
\]
Finally, $\mathrm{EFG} = 0$ if and only if every pairwise difference is zero, which
holds exactly when $\mu_g = \mu_{g'}$ for all $g,g' \in G$. This proves boundedness,
symmetry, and the zero-iff-equal property.

\subsection{Proof of Proposition 2 (Subgroup-scarcity variance effect)}
\label{sec:proof-prop2}

Fix a group $g$ and assume $\{M_i\}_{i \in D_g}$ are i.i.d.\ with variance
$\sigma_g^2$. Then the subgroup mean is
\[
\mu_g = \frac{1}{|D_g|} \sum_{i \in D_g} M_i.
\]
By the variance of the sample mean,
\begin{equation*}
\begin{aligned}
\mathrm{Var}(\mu_g)
&= \mathrm{Var}\left(\frac{1}{|D_g|} \sum_{i \in D_g} M_i\right) \\
&= \frac{1}{|D_g|^2} \sum_{i \in D_g} \mathrm{Var}(M_i) \\
&= \frac{\sigma_g^2}{|D_g|}.
\end{aligned}
\end{equation*}
Thus, the variance decays only at rate $1/|D_g|$. When $|D_g|$ is small,
the subgroup mean is intrinsically noisier. Since
\[
\mathrm{EFG} = \max_{g,g' \in G} \big|\mu_g - \mu_{g'}\big|,
\]
its estimate becomes more sensitive to low-support groups, especially under
long-tailed subgroup distributions. The same reasoning applies to Cohen’s $d$,
since both the numerator and pooled variance become unstable when one subgroup
has limited support. This variance law provides the analytical basis for using
bootstrap confidence intervals and cautious interpretation of sparse-group fairness
estimates.

\subsection{Formal argument for Observation 1 (Partial orthogonality)}
\label{sec:proof-obs1}

Let $F_{\text{pred}}(\theta)$ denote a predictive objective, such as negative
macro-F1 or cross-entropy risk, and let $F_{\text{exp}}(\theta) = \mathrm{EFG}(\theta)$
denote the explanation-fairness objective. Both are differentiable in a neighborhood
of interest.

If the two objectives were perfectly aligned, one would expect the gradients to be
proportional, i.e., there exists a scalar $c > 0$ such that
\[
\nabla_\theta F_{\text{pred}}(\theta) = c \,\nabla_\theta F_{\text{exp}}(\theta)
\quad \text{for all } \theta.
\]
In general, this does not hold. The predictive objective $F_{\text{pred}}(\theta)$
is driven primarily by class-separating features and outcome-level distributions,
whereas the explanation-fairness objective $F_{\text{exp}}(\theta)$ depends on
subgroup-conditioned attribution geometry induced by $A_\theta$, rationale masks
$r_i$, and group-specific token interactions. Attributions are functions of
gradients or coalition marginals,
\[
s_i = A_\theta(x_i),
\]
and subgroup fairness depends on the distribution of
\[
M_i = M(s_i, r_i)
\]
within each $D_g$. The mapping $\theta \mapsto F_{\text{pred}}(\theta)$ is determined
by prediction errors and label distribution, while $\theta \mapsto F_{\text{exp}}(\theta)$
is determined by how attribution scores align with rationales within and across
groups. These two maps are structurally distinct and, for generic architectures and
data, their level sets intersect transversely rather than coinciding. As a result,
one can find directions in parameter space along which $F_{\text{pred}}(\theta)$
decreases while $F_{\text{exp}}(\theta)$ remains unchanged or increases. Therefore,
the gradients are not proportional in general, and local descent for prediction need
not automatically reduce explanation fairness gaps. This formalizes the partial
orthogonality claim stated in the main text.

\subsection{Proof sketch for Proposition 3 (Conditional tradeoff under fairness regularization)}
\label{sec:proof-prop3}

Assume $L_{\text{task}}$, $L_{\text{rat}}$, and $L_{\text{adv}}$ are differentiable,
and that demographic attributes are correlated with some task-relevant features in
the representation $h_\theta(x)$.

Consider the objective
\begin{equation*}
\begin{aligned}
J(\theta,\phi,\lambda_{\text{adv}}) 
&= L_{\text{task}}(\theta) + \lambda_{\text{rat}} L_{\text{rat}}(\theta) \\
&\quad + \lambda_{\text{adv}} L_{\text{adv}}(\theta,\phi).
\end{aligned}
\end{equation*}
Differentiating with respect to $\theta$ gives
\[
\nabla_\theta J 
= \nabla_\theta L_{\text{task}} 
+ \lambda_{\text{rat}} \nabla_\theta L_{\text{rat}} 
+ \lambda_{\text{adv}} \nabla_\theta L_{\text{adv}}.
\]
As $\lambda_{\text{adv}}$ increases, the adversarial term dominates whenever
\[
\lambda_{\text{adv}} \big\|\nabla_\theta L_{\text{adv}}\big\| 
\gg \big\|\nabla_\theta L_{\text{task}} + \lambda_{\text{rat}} \nabla_\theta L_{\text{rat}}\big\|.
\]
In that regime, optimization strongly suppresses group-predictive information in the
shared representation $h_\theta(x)$. If some of that information is entangled with
task-relevant semantics, then suppressing it also attenuates useful classification
signal. Hence, for sufficiently large $\lambda_{\text{adv}}$, it is possible to reduce
$\mathrm{EFG}$ while degrading predictive utility, yielding a non-trivial fairness–utility
Pareto frontier rather than monotone improvement in all metrics. This behavior is
observed empirically in our $\lambda_{\text{adv}}$-sweep and matches prior analyses of
adversarial fairness learning and explanation-based mitigation. We therefore present
Proposition~3 as a conditional optimization intuition rather than a universal theorem.

\subsection{Lemma and formal argument for Observation 2 (Explainer-level ranking stability)}
\label{sec:proof-obs2}

\textbf{Lemma (Monotonic transformation preserves ranking).}
Let $u: \mathcal{X} \to \mathbb{R}$ be any scoring function and let
$f: \mathbb{R} \to \mathbb{R}$ be a strictly increasing (positive monotonic)
function. For any $x,y \in \mathcal{X}$,
\[
u(x) > u(y) \;\Rightarrow\; f(u(x)) > f(u(y)).
\]
Thus, $f \circ u$ induces the same ranking over $\mathcal{X}$ as $u$.

\textit{Proof.}
By strict monotonicity, $x > y$ implies $f(x) > f(y)$ for all $x,y \in \mathbb{R}$.
Setting $x = u(x)$ and $y = u(y)$ yields $u(x) > u(y) \Rightarrow f(u(x)) > f(u(y))$,
so the order of items induced by $u$ is preserved under $f \circ u$. \qed

\medskip

\textbf{Application to IG vs SHAP ranking.}
Let $A_\theta^{\text{IG}}$ and $A_\theta^{\text{SHAP}}$ be two attribution operators,
and define normalized non-negative scores
\begin{equation*}
\begin{aligned}
s_i^{\text{IG}} &= \mathrm{Norm}\big(\max(A_\theta^{\text{IG}}(x_i), 0)\big), \\
s_i^{\text{SHAP}} &= \mathrm{Norm}\big(\max(A_\theta^{\text{SHAP}}(x_i), 0)\big).
\end{aligned}
\end{equation*}
Suppose there exists a latent importance ranking $u_i$ and two strictly increasing
transformations $f_{\text{IG}}, f_{\text{SHAP}}$ such that, coordinatewise,
\[
s_i^{\text{IG}} = f_{\text{IG}}(u_i),
\quad
s_i^{\text{SHAP}} = f_{\text{SHAP}}(u_i).
\]
Then by the lemma, both IG- and SHAP-based plausibility functionals
$M(s_i^{\text{IG}}, r_i)$ and $M(s_i^{\text{SHAP}}, r_i)$ preserve the ranking over
models induced by the latent importance scores, even though absolute values and
metric scales may differ.

In practice, IG can be viewed as a path integral over gradients and SHAP as a
coalition-value estimator, so they are not guaranteed to be strict monotonic
transforms of each other. Nonetheless, our empirical results show that IG and
SHAP agree on the comparative ordering of models (BERT $<$ RoBERTa $<$ ADV-RoBERTa)
while differing in absolute plausibility scale and subgroup spread. This justifies
interpreting cross-explainer agreement primarily in terms of ranking stability
(which is preserved under monotonic transforms), rather than identical attribution
magnitudes, and motivates reporting both explainers when analyzing explanation
fairness.
\section{Implementation and Reproducibility Details}
\label{sec:implementation}

\subsection*{D.1 Baseline Justification}

We evaluate BERT, RoBERTa, and ADV-RoBERTa rather than DeBERTa, HateBERT, or instruction-tuned large language models for three reasons. First, BERT provides a canonical transformer baseline whose classification and attribution behavior are well studied in the fairness and interpretability literature. Second, RoBERTa is the strongest general-purpose encoder in the same parameter regime and is widely used in NLP fairness research, making it the most appropriate comparative backbone for HateXplain. Third, ADV-RoBERTa isolates the effect of adversarial demographic debiasing by keeping the architecture fixed while modifying only the training objective.

We exclude DeBERTa and HateBERT because they introduce architectural or pretraining confounds that would make it difficult to attribute any improvement in explanation fairness specifically to debiasing. We also exclude instruction-tuned decoder-only LLMs because our plausibility framework relies on token-level gradients or coalition-based attributions over the input embedding space, which are not directly comparable to the encoder-based setting studied here without substantial methodological changes.

\subsection*{D.2 Detailed Model Architectures}

BERT \cite{devlin2019bert} is a 12-layer bidirectional transformer encoder with hidden dimension $d=768$ and 12 self-attention heads. Given input $x$, the encoder produces hidden states $H=E_\theta(x) \in \mathbb{R}^{T \times d}$, and the [CLS] representation is projected through a linear classification head:
\[
\hat{y} = \text{softmax}(W_C h_{\text{CLS}} + b_C), \quad W_C \in \mathbb{R}^{|Y| \times d}.
\]

RoBERTa \cite{liu2019roberta} uses the identical architecture but is pretrained with dynamic masking, larger batch sizes, longer training schedules, and removal of the next-sentence prediction objective. These modifications produce substantially stronger contextual representations. We use \texttt{roberta-base} with the same classification head structure.

\begin{figure*}[t]
    \centering
    \includegraphics[width=\textwidth]{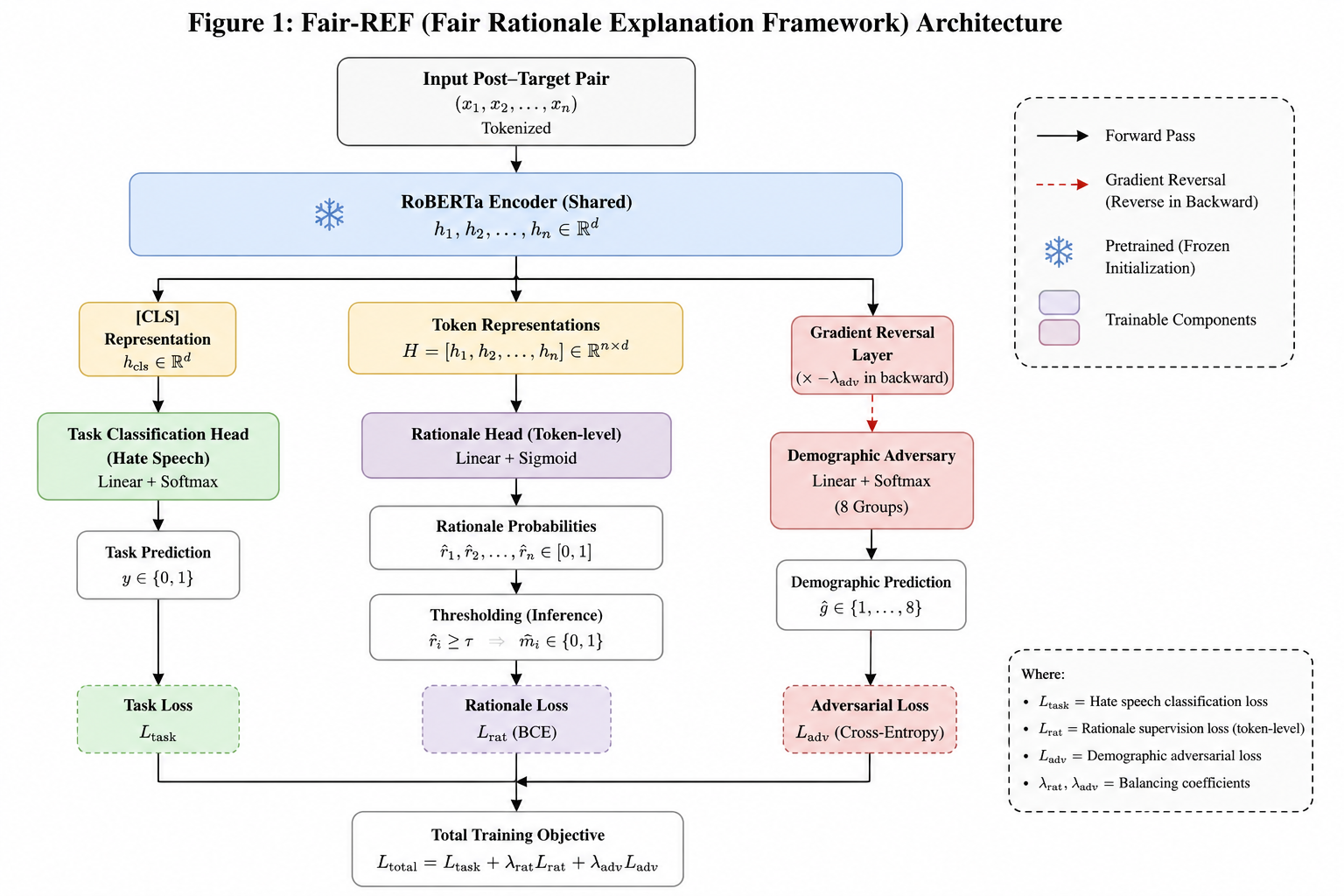}
    \caption{\textbf{Fair-REF (Fair Rationale Explanation Framework) architecture.} A shared RoBERTa encoder feeds three task-specific branches: (i) a hate speech classification head optimized with the task loss ($L_{\mathrm{task}}$), (ii) a token-level rationale prediction head trained using binary cross-entropy rationale supervision ($L_{\mathrm{rat}}$), and (iii) a gradient-reversal demographic adversary trained with adversarial loss ($L_{\mathrm{adv}}$). The three objectives are jointly optimized according to $L_{\mathrm{total}} = L_{\mathrm{task}} + \lambda_{\mathrm{rat}}L_{\mathrm{rat}} + \lambda_{\mathrm{adv}}L_{\mathrm{adv}}$, encouraging accurate prediction, faithful rationale generation, and improved explanation fairness across demographic groups.}
    \label{fig:architecture}
\end{figure*}

ADV-RoBERTa extends RoBERTa with adversarial demographic debiasing and explicit rationale supervision. The architecture consists of a shared encoder $E_\theta$, a task classification head $C_\psi$, a demographic adversary $A_\phi$, and a rationale prediction head $R_\xi$. The rationale head allows the model to be directly supervised with human-annotated rationale masks, creating an explicit alignment signal between learned representations and human reasoning. This design is motivated by the adversarial invariance learning paradigm of \cite{zhang2018mitigating} and the rationale supervision approach established in the HateXplain benchmark \cite{mathew2021hatexplain}. Figure~\ref{fig:architecture} illustrates the complete Fair-REF architecture.
\subsection*{D.3 Forward Pass and Loss Construction}

Given input $x$, the encoder produces the contextual hidden states and the [CLS] representation:
\[
H = E_\theta(x), \quad h = H_{\text{[CLS]}} \in \mathbb{R}^d.
\]
The task head predicts the class distribution as $\hat{y}=C_\psi(h)=\text{softmax}(W_C h+b_C)$. The adversary predicts the demographic target group from the gradient-reversed representation:
\[
\hat{g}=A_\phi(\mathrm{GRL}_\lambda(h))=\text{softmax}(W_A h+b_A).
\]
The rationale prediction head operates over the full token sequence to produce per-token rationality scores:
\[
\hat{r}_t=\sigma(R_\xi(H_t)) \in [0,1], \quad \forall t \in \{1,\dots,T\},
\]
where $\sigma(\cdot)$ is the sigmoid function.

The task loss is standard categorical cross-entropy:
\[
L_{\text{task}} = -\sum_{k \in Y} I[y=k]\log \hat{y}_k.
\]
The adversarial loss is categorical cross-entropy over the demographic prediction:
\[
L_{\text{adv}} = -\sum_{k \in G} I[g=k]\log \hat{g}_k.
\]
The rationale supervision loss is token-level binary cross-entropy against the human rationale mask:
\[
L_{\text{rat}} = -\frac{1}{T}\sum_{t=1}^T \big[r_t\log \hat{r}_t + (1-r_t)\log(1-\hat{r}_t)\big].
\]
$L_{\text{rat}}$ provides a direct supervision signal over which tokens the model should treat as salient, creating explicit alignment between the model’s token-level rationality scores and the human-annotated evidence spans. Only instances with valid rationale annotations $(\sum_t r_t > 0)$ contribute to this term.

The full optimization objective for ADV-RoBERTa is
\[
L_{\text{total}} = L_{\text{task}} + \lambda_{\text{adv}} L_{\text{adv}} + \lambda_{\text{rat}} L_{\text{rat}},
\]
where $\lambda_{\text{adv}} \in \mathbb{R}_{>0}$ controls adversarial suppression strength and $\lambda_{\text{rat}} \in \mathbb{R}_{>0}$ controls the rationale alignment weight. We set $\lambda_{\text{adv}}=0.5$ and $\lambda_{\text{rat}}=0.2$ in all primary experiments, based on validation macro-F1 and Plausibility-EFG.

\subsection*{D.4 Gradient Reversal and Fairness Ablation}

The Gradient Reversal Layer (GRL; Ganin et al., 2016) acts as the identity during the forward pass,
\[
\mathrm{GRL}_\lambda(h)=h,
\]
and multiplies the gradient by $-\lambda_{\text{adv}}$ during backpropagation:
\[
\frac{\partial \mathrm{GRL}_\lambda(h)}{\partial h}=-\lambda_{\text{adv}} \cdot I.
\]
This coupling implements the inner maximization over $\phi$ implicitly, without requiring alternating optimization steps. The full min–max objective can be written as
\[
\min_{\theta,\psi,\xi} \max_\phi \; L_{\text{task}}+\lambda_{\text{rat}}L_{\text{rat}}-\lambda_{\text{adv}}L_{\text{adv}}.
\]
The encoder is thereby trained simultaneously to minimize classification loss, minimize rationale prediction error, and maximize the demographic adversary’s prediction error, producing representations that are task-discriminative, rationale-aligned, and demographically invariant.

To characterize the fairness–performance tradeoff, we introduce $\lambda_{\text{fair}}$ as the ablation variable for adversarial debiasing strength:
\[
L_{\text{total}} = L_{\text{task}} + \lambda_{\text{fair}} L_{\text{adv}} + \lambda_{\text{rat}} L_{\text{rat}},
\]
where $\lambda_{\text{fair}} \in \{0.0, 0.1, 0.3, 0.5, 1.0\}$ and $\lambda_{\text{rat}}=0.2$ is held fixed. When $\lambda_{\text{fair}}=0$, the adversarial branch is fully disabled and the model reduces to a RoBERTa encoder with rationale supervision only. As $\lambda_{\text{fair}}$ increases, the model places progressively stronger pressure on the encoder to suppress demographic information from the hidden state, yielding the fairness–utility Pareto frontier.

\subsection*{D.5 Training Dynamics and Stability}

Adversarial training with gradient reversal can be unstable early in training, so we use five safeguards: linear warmup of $\lambda_{\text{adv}}$ over the first 10\% of steps, gradient clipping at 1.0, dropout rate 0.1 on the [CLS] representation, checkpoint selection by validation macro-F1, and early stopping with patience on validation macro-F1. These choices help stabilize the inner adversarial game while preserving classification performance.

\subsection*{D.6 Attribution Details}

Integrated Gradients (IG; Sundararajan et al., 2017) computes token attributions by integrating the gradient of the target-class output along a straight-line path from a zero-vector baseline $x'=0$ to the actual input $x$. The attribution for token $t$ is
\begin{equation*}
\begin{aligned}
\mathrm{IG}_t(x) &= (x_t-x'_t)\int_0^1 \frac{\partial F(x'+\alpha(x-x'))}{\partial x_t} \,d\alpha \\
&\approx (x_t-x'_t) \cdot \frac{1}{50}\sum_{k=1}^{50} \frac{\partial F\big(x'+\tfrac{k}{50}(x-x')\big)}{\partial x_t}.
\end{aligned}
\end{equation*}
Attributions are computed at the word embedding level, reduced to scalar scores via the $\ell_2$-norm over the embedding dimension, normalized per instance to $[0,1]$, and clipped to zero for negative values, since negative attributions indicate features that suppress the predicted class and are not evidence for the rationale.

SHAP \cite{lundberg2017unified} estimates token importance via Shapley values. For token $t$,
\[
\phi_t = \sum_{S \subseteq N \setminus \{t\}} \frac{|S|!(|N|-|S|-1)!}{|N|!}\big[f(S \cup \{t\}) - f(S)\big].
\]
We use a text masker with a maximum evaluation budget of 256 calls per sample, retaining only non-negative values and normalizing per instance to $[0,1]$. IG and SHAP are then fed into the plausibility and fairness metrics defined in Section~4, providing two complementary views of attribution behavior.

\subsection*{D.7 Plausibility and Fairness Metrics}

For completeness, we summarize the metrics used; full definitions are in Section~4. Let $s_i \in [0,1]^{T_i}$ be the attribution vector and $r_i \in \{0,1\}^{T_i}$ the human rationale mask. AUPRC evaluates the ranking quality of $s_i$ against $r_i$, treating rationale tokens as positives:
\[
\text{AUPRC}(s_i, r_i) = \int_0^1 P(\tau) \,dR(\tau).
\]
IOU-F1 binarizes attribution scores at the per-instance median $\tau_i = \text{median}(s_i)$:
\begin{equation*}
\begin{aligned}
\hat{r}_{it} &= I[s_{it} \ge \tau_i], \\
\text{IOU-F1}(i) &= \frac{|\hat{r}_i \cap r_i|}{|\hat{r}_i \cup r_i|}.
\end{aligned}
\end{equation*}
For demographic group $g \in G$, group-wise mean plausibility is
\[
\mu_g = \frac{1}{|D_g|} \sum_{i \in D_g} M_i,
\]
and Plausibility-EFG is the worst-case inter-group disparity:
\[
\text{EFG} = \max_{g_i, g_j \in G} |\mu_{g_i} - \mu_{g_j}|.
\]
We report Kruskal–Wallis $H$-statistics, Cohen’s $d$ between best and worst groups, and bootstrap 95\% confidence intervals over $\mu_g$ to quantify statistical and practical subgroup disparities.

\subsection*{D.8 Training and Hardware Configuration}

All models are trained under identical preprocessing and optimization settings on a single NVIDIA A100 40GB GPU using PyTorch, HuggingFace Transformers, Captum, and the SHAP library. Each post is tokenized with the backbone-specific HuggingFace tokenizer, truncated or zero-padded to a maximum sequence length of 128 tokens, and passed with an attention mask. The optimization configuration is:

\begin{center}
\begin{tabular}{ll}
\toprule
Hyperparameter & Value \\
\midrule
Optimizer & AdamW \\
Learning rate & $2 \times 10^{-5}$ \\
Batch size & 32 \\
Max sequence length & 128 \\
Weight decay & 0.01 \\
Dropout (CLS) & 0.10 \\
Gradient clipping & 1.0 \\
Warmup proportion & 10\% of steps \\
Epochs (BERT/RoBERTa) & 10 \\
Epochs (ADV-RoBERTa) & 20 \\
Random seed & 42 \\
GPU & NVIDIA A100 40GB \\
\bottomrule
\end{tabular}
\end{center}

A linear warmup schedule runs over the first 10\% of total optimization steps followed by linear decay. ADV-RoBERTa uses adversary warmup over the first 10\% of steps to allow the adversarial objective to stabilize. The best checkpoint in all cases is selected by validation macro-F1. The fairness coefficient is swept over $\lambda_{\text{fair}} \in \{0.0, 0.1, 0.3, 0.5, 1.0\}$ to study the fairness–performance tradeoff.

\subsection*{D.9 Dataset Construction Pipeline}

To make the evaluation protocol reproducible, we explicitly describe the transformation from the raw HateXplain posts to the fairness-evaluation set. Starting from the 20,148-post HateXplain corpus, we apply the following deterministic pipeline:
\begin{enumerate}
    \item \textbf{Annotator expansion ($20{,}148 \rightarrow 60{,}444$).} Each post is expanded to one row per annotator; HateXplain provides approximately three annotator judgments per post, yielding 60,444 annotation-level rows.
    \item \textbf{Target-group explosion ($60{,}444 \rightarrow 53{,}161$).} Each annotation’s target-group list is exploded to one row per $(\text{annotation}, \text{group})$. Annotations with no target group (chiefly normal-labeled posts) are dropped, which is why this step nets to a smaller count despite the explosion.
    \item \textbf{Rationale and group filtering ($53{,}161 \rightarrow 27{,}566$).} The exploded set is filtered to the eight canonical identity groups with a non-empty rationale mask, giving the final fairness-evaluation set used for AUPRC/IOU-F1 and Plausibility-EFG.
\end{enumerate}
Posts with multiple annotated targets therefore appear in multiple group rows and contribute shared attribution scores to more than one subgroup’s plausibility estimate. This design partially addresses multi-target posts but does not construct explicit intersectional categories; the limitations of this single-label taxonomy and the induced dependence structure are discussed in the main Limitations section.

\subsection*{D.10 Orthogonality and Mitigation Framing}

The central empirical finding of this work is that prediction fairness and explanation fairness are only partially orthogonal: they are related, but improving one does not guarantee improvement in the other. In practice, stronger accuracy or debiasing can reduce Plausibility-EFG, but it does not eliminate explanation disparity, so explanation fairness must be treated as a first-class optimization target rather than a by-product of classification performance. The Fair-REF framework (ADV-RoBERTa with rationale supervision and adversarial debiasing) is therefore interpreted as a bounded mitigation strategy that directly lowers Plausibility-EFG through explicit rationale alignment and adversarial invariance, while higher $\lambda_{\text{fair}}$ values introduce a measurable performance cost. This reflects a Pareto tradeoff between explanation fairness and predictive utility, consistent with the conditional tradeoff intuition formalized in Proposition~3.

\section{Ethical Considerations}

This work involves hate speech detection, identity-linked subgroup analysis, and explanation fairness, so the primary ethical concern is the potential for harm through misclassification or misinterpretation. False positives can suppress legitimate speech, while false negatives can fail to flag abusive content, so the system should be framed as a research tool rather than an automated moderation authority. The fairness analysis is intended to reduce disparate treatment across demographic groups, but subgroup labels themselves are sensitive and should be handled with care.

A second concern is that human rationales are not ground truth in an absolute sense; they are annotator-provided evidence spans that can reflect disagreement, ambiguity, or cultural context. For that reason, plausibility metrics should be interpreted as agreement with annotations rather than proof of explanation correctness. The use of subgroup explosion also increases the risk of over-interpreting small-group behavior, so results should be discussed with attention to sample imbalance and uncertainty. Finally, any deployment context should include human oversight, monitoring for bias drift, and clear limitations on using the model for high-stakes moderation decisions.

\end{document}